\documentclass{article}

\usepackage[preprint]{neurips_2023}

\usepackage{amsmath}
\usepackage{pdflscape}

\usepackage[dvipsnames, svgnames, x11names]{xcolor}
\usepackage[utf8]{inputenc} %
\usepackage[T1]{fontenc}    %
\usepackage{hyperref}       %
\usepackage{url}            %
\usepackage{booktabs}       %
\usepackage{amsfonts}       %
\usepackage{nicefrac}       %
\usepackage{microtype}      %
\usepackage{xcolor}         %
\usepackage{multirow}
\usepackage{CJKutf8}
\usepackage{colortbl}
\usepackage{svg}
\usepackage{amssymb}%
\usepackage{pifont}%

\usepackage{graphicx}

\usepackage{natbib}
\setcitestyle{numbers,square}

\definecolor{'deep1'}{HTML}{C5E6F8} 
\definecolor{'shallow1'}{HTML}{E4F3FC} 
\definecolor{'deep2'}{HTML}{E5F5B7} 
\definecolor{'shallow2'}{HTML}{F3FADF} 

\definecolor{'deep3'}{HTML}{FFE5C6} 
\definecolor{'shallow3'}{HTML}{FFF2E3}

\usepackage{todonotes}
\makeatletter
\newcommand*\iftodonotes{\if@todonotes@disabled\expandafter\@secondoftwo\else\expandafter\@firstoftwo\fi}  %
\makeatother

\title{ EnviroExam: Benchmarking Environmental Science Knowledge of Large Language Models}

\author{%
  Yu Huang$^1$, Liang Guo$^{1\ddagger}$, Wanqian Guo$^{1\ddagger}$, Zhe Tao$^{1\dagger}$,Yang Lv$^{1\dagger}$\\
  \textbf{ Zhihao Sun$^{1\dagger}$, Dongfang Zhao$^1$} \\
  $^1$School of Environment, Harbin Institute of Technology \\
  \hspace{-0.25cm}$^{\dagger}$ Equal Contribution  \ \ $^{\ddagger}$ Corresponding Author\\
  \url{https://enviroexam.enviroscientist.cn}
}

\begin{document}

\maketitle

\begin{abstract}
In the field of environmental science, it is crucial to have robust evaluation metrics for large language models to ensure their efficacy and accuracy. We propose EnviroExam, a comprehensive evaluation method designed to assess the knowledge of large language models in the field of environmental science. EnviroExam is based on the curricula of top international universities, covering undergraduate, master's, and doctoral courses, and includes 936 questions across 42 core courses. By conducting 0-shot and 5-shot tests on 31 open-source large language models, EnviroExam reveals the performance differences among these models in the domain of environmental science and provides detailed evaluation standards. The results show that 61.3\% of the models passed the 5-shot tests, while 48.39\% passed the 0-shot tests. By introducing the coefficient of variation as an indicator, we evaluate the performance of mainstream open-source large language models in environmental science from multiple perspectives, providing effective criteria for selecting and fine-tuning language models in this field. Future research will involve constructing more domain-specific test sets using specialized environmental science textbooks to further enhance the accuracy and specificity of the evaluation.

\end{abstract}

\section{Introduction}
Since the invention of the Transformer architecture in 2017\cite{vaswani2023attention}, the development of large language models has accelerated. In this context, the emergence of both closed-source models like ChatGPT\cite{chatgpt2023applications} and Claude\cite{claude2023}, and open-source models like Llama\cite{touvron2023llama} and Qwen\cite{bai2023qwen}, has addressed the challenge of extracting valuable information from vast knowledge bases and, in some cases, surpassed human intelligence. To effectively leverage large language models in environmental science, establishing evaluation standards has become essential\cite{doi:10.1021/acs.est.3c01818,AGATHOKLEOUS2023164154,Boiko2023}.

Researchers have established knowledge assessment tests for general large language models, with notable test sets including Ceval\cite{huang2023ceval}, MMLU\cite{hendrycks2021measuring}, and HumanEval\cite{chen2021evaluating}. These tests evaluate model capabilities across various disciplines and rank them, providing a clear understanding of their strengths and weaknesses for developers and users. However, due to the strong specialization and unique characteristics of vertical fields, these general tests often fail to cover the content of environmental science, thereby limiting the development of specialized language models in this domain.

The establishment of evaluation standards for vertical domain language models, such as Lawbench\cite{fei2023lawbench,LaWGPT2023} and Medbench\cite{cai2023medbench}, demonstrates the feasibility and potential of creating such standards in specific fields. Lawbench evaluates 52 mainstream models from the perspectives of memory, understanding, and application, achieving notable results. Medbench, through a comprehensive medical competency test set, has conducted thorough evaluations of 113 models, yielding highly credible outcomes. These successes in the legal and medical fields suggest that it is feasible to establish reliable and effective evaluation standards for large language models in environmental science.

\begin{figure}[t]
    \centering
    \includegraphics[width=\columnwidth]{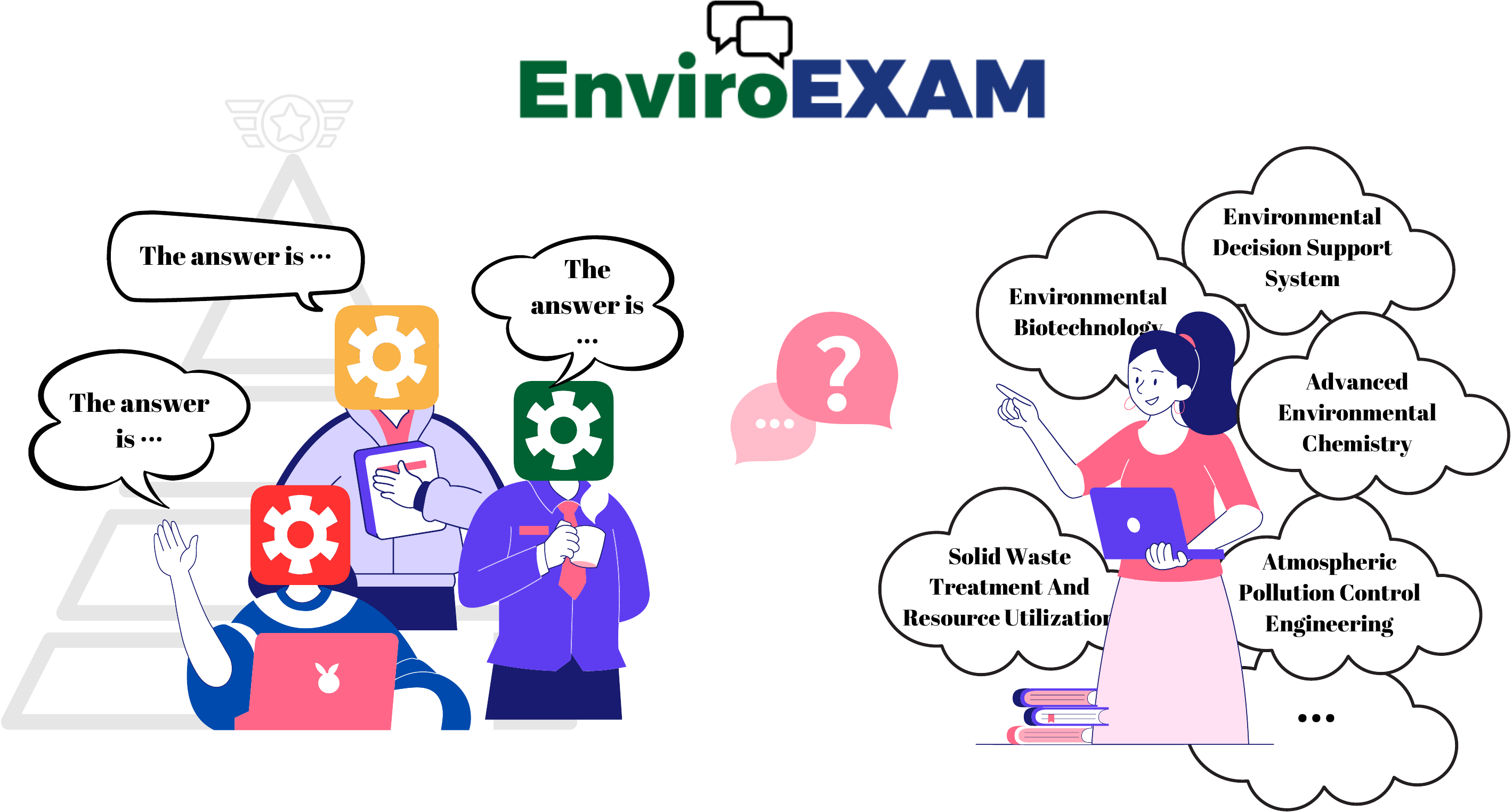}
    \caption{Overview diagram of EnviroExam. Treat different large language models as different students, assess them and rank them accordingly.}
    \label{fig:exam}
\end{figure}

Considering these points, we established EnviroExam\ref{fig:exam}, a comprehensive evaluation designed to assess large language models' performance in environmental science. Based on the curriculum of top international universities and covering undergraduate, master's, and doctoral courses, EnviroExam includes 936 questions across 42 core environmental science courses.Detailed courses information and the number of questions can be found in Appendix \hyperref[appendixA]{A}.

\section{The EnviroExam Evaluation Suite}
\paragraph{Design philosophy}
The design concept of EnviroExam is inspired by the core course assessments for college students. We believe that general artificial intelligence represents high school or undergraduate students, while vertical domain large language models represent graduate or doctoral students. Under this concept, we view different large language models as different students, utilizing university assessment methods to evaluate and rank them. 

\begin{figure}
    \centering
    \includegraphics[width=1\linewidth]{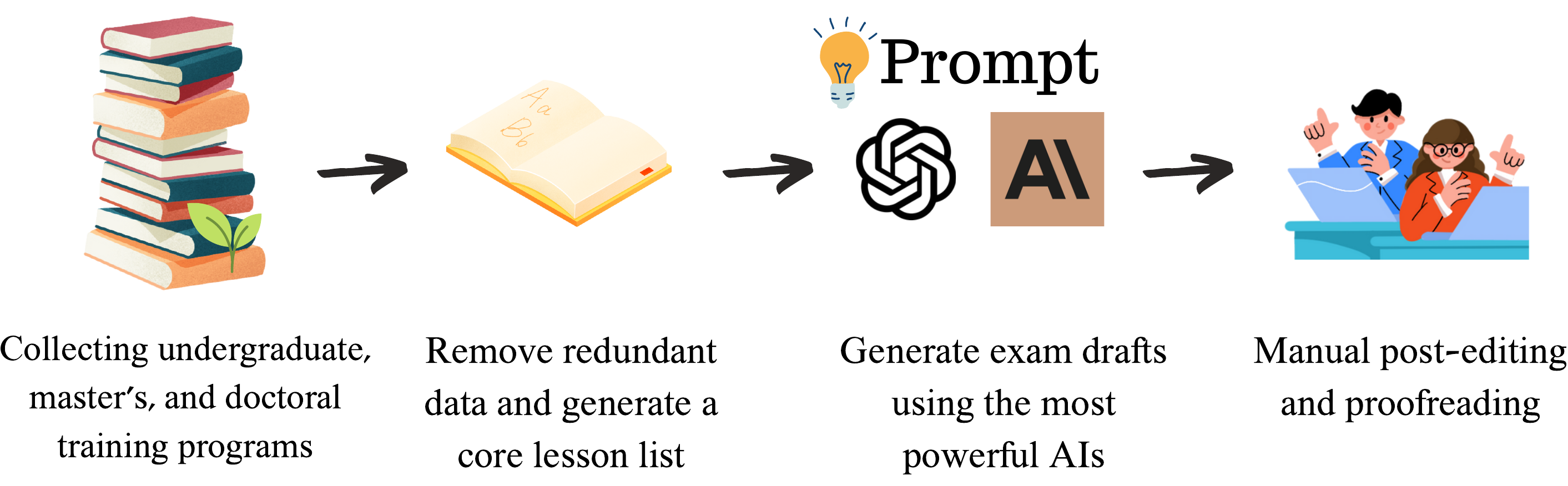}
    \caption{Data collection and process workflows}
    \label{fig:work_flow}
\end{figure}

\paragraph{Data collection and process}
EnviroExam focuses on 42 core courses from the environmental science curriculum at Harbin Institute of Technology\footnote{We use Harbin Institute of Technology as an example because its Environmental Science and Engineering program received an A+ rating in the fourth round of university evaluations\cite{cdgdc2023}. (An A+ rating in university discipline evaluations signifies that the program is among the top in its field nationwide.)}, after excluding general, duplicate, and practical courses from a total of 141 courses across undergraduate, master's, and doctoral programs. For these 42 courses, initial draft questions were generated using GPT-4\cite{openai2024gpt4} and Claude\cite{kevian2024capabilities}, combined with customized prompts(For details, see  Appendix \hyperref[appendixB]{B}). These drafts were then refined and proofread manually, resulting in a total of 1,290 multiple-choice questions. After final proofreading and refinement, 936 valid questions remained(Example questions can be found in Appendix \hyperref[appendixC]{C}), which were divided into two datasets: 210 questions for the development set (dev) and 726 questions for the test set (test). The data processing workflow is shown in the Figure\ref{fig:work_flow}.

\paragraph{Scoring Method}

EnviroExam uses accuracy as the basis for scoring each subject's questions and employs a comprehensive metric when calculating the total score. The formula derivation process is as follows: \\
1. Calculation of the average score $M$: For each large language model, compute the arithmetic mean of all its test scores (accuracy):\\
\[
M = \frac{1}{n} \sum_{i=1}^{n} s_i
\]
Where $s_i$ is the score of a large language model on a specific test, and $n$ is the total number of tests.\\
2. Calculate the standard deviation: Compute the standard deviation of all test scores relative to the mean:
\[
\sigma = \sqrt{\frac{1}{n} \sum_{i=1}^{n} (s_i - 1)^2}
\]
3. Calculate the coefficient of variation (CV): The coefficient of variation is the ratio of the standard deviation to the mean and is used to measure the relative dispersion of the scores:
\[
\text{CV} = \frac{\sigma}{M}
\]
4. Calculate the original composite index \( I \): The original composite index \( I \) is defined as a function of the mean score and the coefficient of variation:
\[
I = 
\left\{
\begin{array}{ll}
M \times (1 - \text{CV}), & 0 \leq \text{CV} \leq 1 \\
\text{model void}^*\footnotemark, & \text{CV} > 1
\end{array}
\right.
\]

\footnotetext{When CV is greater than 1, it indicates that the relative variability of the data is very high, and the mean can no longer effectively represent the central tendency of the data\cite{10.3389/fams.2019.00043}.}
\paragraph{Models}
We used 31 different open-source large language models with varying parameter sizes and sources for this generation. The diversity of these large language models allows for more effective testing of their performance in the field of environmental science. The parameter sizes are shown in the Table\ref{tab:models}, and detailed descriptions of the models can be found in Appendix \hyperref[appendixD]{D}.

\begin{table}[h]
\centering
\begin{tabular}{llll}
\toprule
Model & Creator & \#Parameters & Access \\
\midrule
baichuan2-13b-chat & Baichuan-ai & 13B & Weights \\
baichuan-13b-chat & Baichuan-ai & 13B & Weights \\
chatglm2-6b & THUDM & 6B & Weights \\
chatglm3-6b & THUDM & 6B & Weights \\
chatglm3-6b-32k & THUDM & 6B & Weights \\
deepseek-7b-chat & Deepseek ai & 7B & Weights \\
deepseek-67b-chat & Deepseek ai & 67B & Weights \\
gemma-7b & Google & 7B & Weights \\
gemma-2b-it & Google & 2B & Weights \\
internlm2-chat-20b & Shanghai AI Laboratory & 20B & Weights \\
internlm2-chat-7b & Shanghai AI Laboratory & 7B & Weights \\
internlm-chat-20b & Shanghai AI Laboratory & 20B & Weights \\
internlm-chat-7b & Shanghai AI Laboratory & 7B & Weights \\
mistral-7b-instruct-v0.1 & Mistralai & 7B & Weights \\
mixtral-8x7b-instruct-v0.1 & Mistralai & 56B & Weights \\
qwen1.5-14b-chat & Alibaba Cloud & 14B & Weights \\
qwen1.5-7b-chat & Alibaba Cloud & 7B & Weights \\
qwen-14b-chat & Alibaba Cloud & 14B & Weights \\
qwen-7b-chat & Alibaba Cloud & 7B & Weights \\
vicuna-13b-v1.5 & LMSYS & 13B & Weights \\
vicuna-7b-v1.5 & LMSYS & 7B & Weights \\
vicuna-7b-v1.5-16k & LMSYS & 7B & Weights \\
bluelm-7b-chat & Vivo & 7B & Weights \\
tigerbot-13b-chat-v2 & TigerResearch & 13B & Weights \\
tigerbot-7b-chat-v3 & TigerResearch & 7B & Weights \\
llama-3-8b-instruct & Meta & 8B & Weights \\
llama-3-70b-instruct & Meta & 70B & Weights \\
llama-2-70b-chat & Meta & 70B & Weights \\
llama-2-13b-chat & Meta & 13B & Weights \\
skywork-13b & Skywork & 13B & Weights \\
yi-6b-chat & 01-ai & 6B & Weights \\
\bottomrule
\end{tabular}
\caption{ Models evaluated in this paper}
\label{tab:models}
\end{table}

\section{Experiment}
\begin{figure}
    \centering
    \includegraphics[width=1\linewidth]{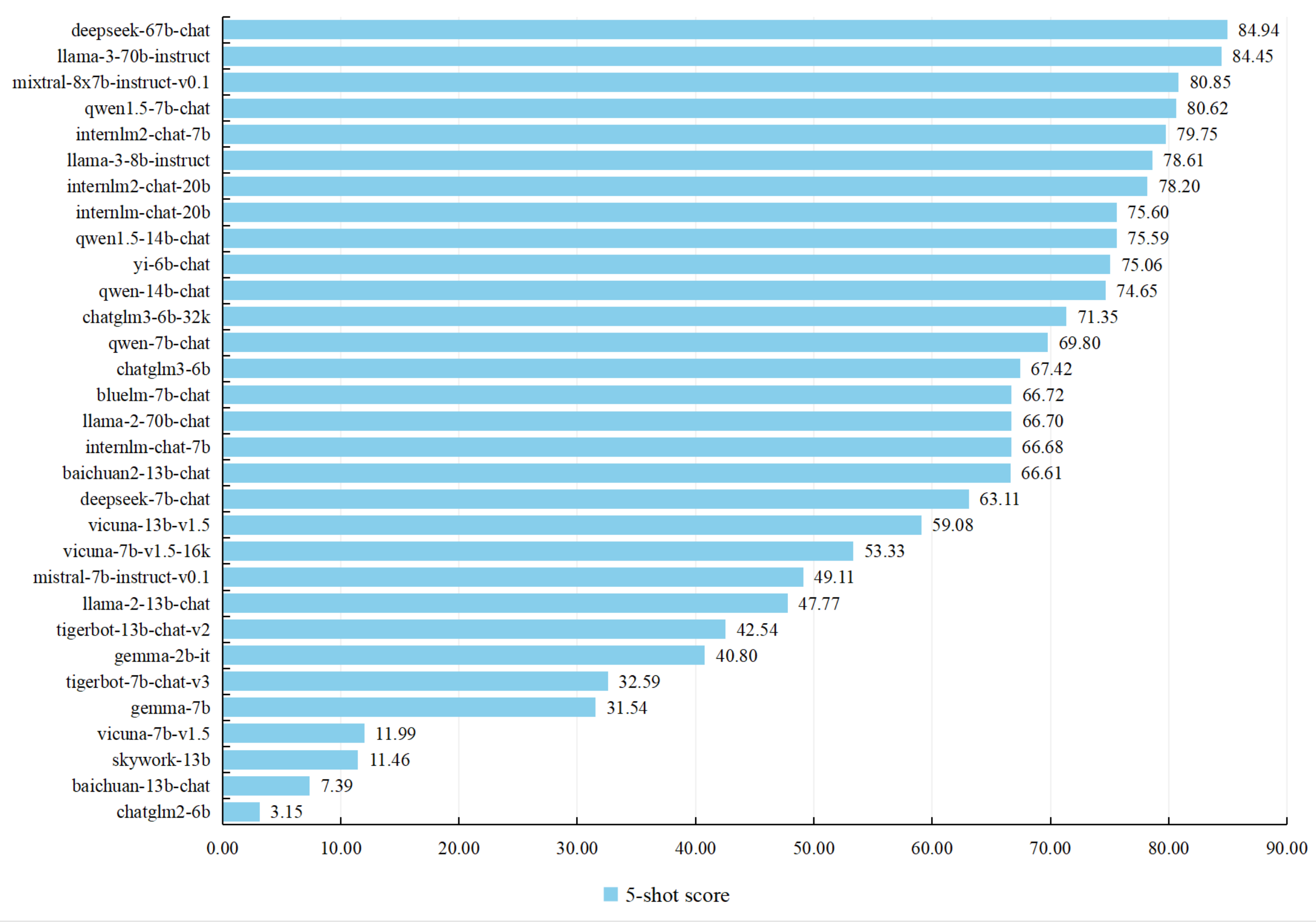}
    \caption{The composite index scores $I$ of 31 LLMs evaluated on EnviroExam(5-shot).}
    \label{fig:5-shot}
\end{figure}

\paragraph{Experiment Setting}
We used OpenCompass\footnote{OpenCompass is an open-source benchmarking platform designed to evaluate and compare the performance of large language models (LLMs) across various tasks.}\cite{OpenCompass2023} (ver 2.1.0) for model evaluation and configured the 31 large language models with the following parameters: max\_out\_len=100, max\_seq\_len=4096, temperature=0.7, and top\_p=0.95. We tested all models using both 0-shot and 5-shot methods.
\paragraph{5-shot score}
In the 5-shot experiment, we first provide the model with 5 annotated examples. These examples cover different types of environmental science questions and include answers to help the model understand the structure and expected format of the responses. Next, we present the model with new test questions and ask it to generate answers based on the previously learned 5 examples. The model's generated answers are recorded and compared with the standard answers to evaluate its performance in understanding and generating new answers. This process aims to test the model's learning and generalization abilities with limited examples.Scores are shown in Figure \ref{fig:5-shot}.

\begin{figure}
    \centering
    \includegraphics[width=1\linewidth]{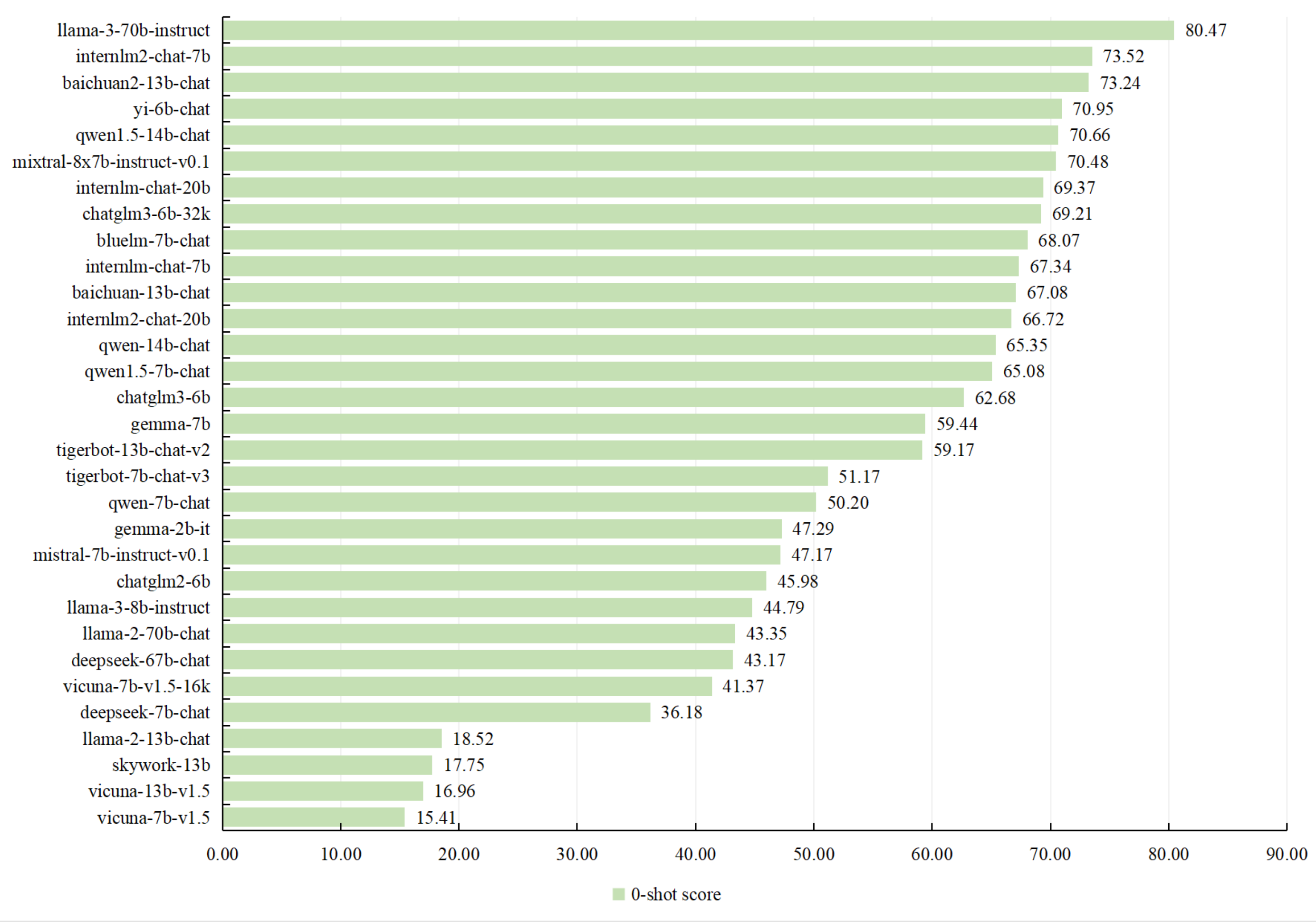}
    \caption{The composite index scores $I$ of 31 LLMs evaluated on EnviroExam(0-shot).}
    \label{fig:0-shot}
\end{figure}

\paragraph{0-shot score}
In the 0-shot experiment, we present the model with test questions without providing any annotated examples beforehand. The model is required to generate answers based solely on its pre-existing knowledge and training. The generated answers are then recorded and compared with the standard answers to evaluate the model's performance in understanding and answering new questions without prior examples. This process aims to test the model's ability to generalize and perform accurately without any specific guidance or context provided through examples.Scores are shown in Figure \ref{fig:0-shot}.

\section{Result}
\paragraph{General comparison}
Based on the overall results of the 5-shot and 0-shot tests, the existing open-source large language models demonstrate relatively good proficiency in the field of environmental science. In the 5-shot tests, where a passing score is considered 60, 19 models passed while 12 models failed, resulting in a pass rate of 61.3\%. In the 0-shot tests, 15 models passed and 16 models failed, yielding a pass rate of 48.39\%. The highest scoring model in the 5-shot tests was DeepSeek-67B-Chat, with a score of 84.94. In the 0-shot tests, the highest scoring model was Llama-3-70B-Instruct, with a score of 80.47.
\paragraph{Does few-shot prompting provide assistance?}
Based on the summarized score results above, we can generate the following radar charts for 5-shot and 0-shot tests. From this radar chart, we can observe that nearly one-third of the large language models did not show an improvement in scores with additional prompts; on the contrary, their scores decreased. This may be because the models have sacrificed few-shot capabilities to enhance their 0-shot performance.
\paragraph{Does chain-of-thought prompting help?}
For most large language models, the use of Chain-of-Thought (COT) prompting is effective and positive. The most significant positive effects were observed in vicuna-13b-v1.5, deepseek-67b-chat, and llama-3-8b-instruct, with scores increasing by 42.12, 41.77, and 33.82 points, respectively. However, for other large language models, the introduction of COT led to a surge in error rates when handling questions. The most significant negative impacts were seen in baichuan-13b-chat, chatglm2-6b, and gemma-7b, with scores decreasing by 59.69, 42.83, and 27.9 points, respectively.
\begin{figure}
    \centering
    \includegraphics[width=1\linewidth]{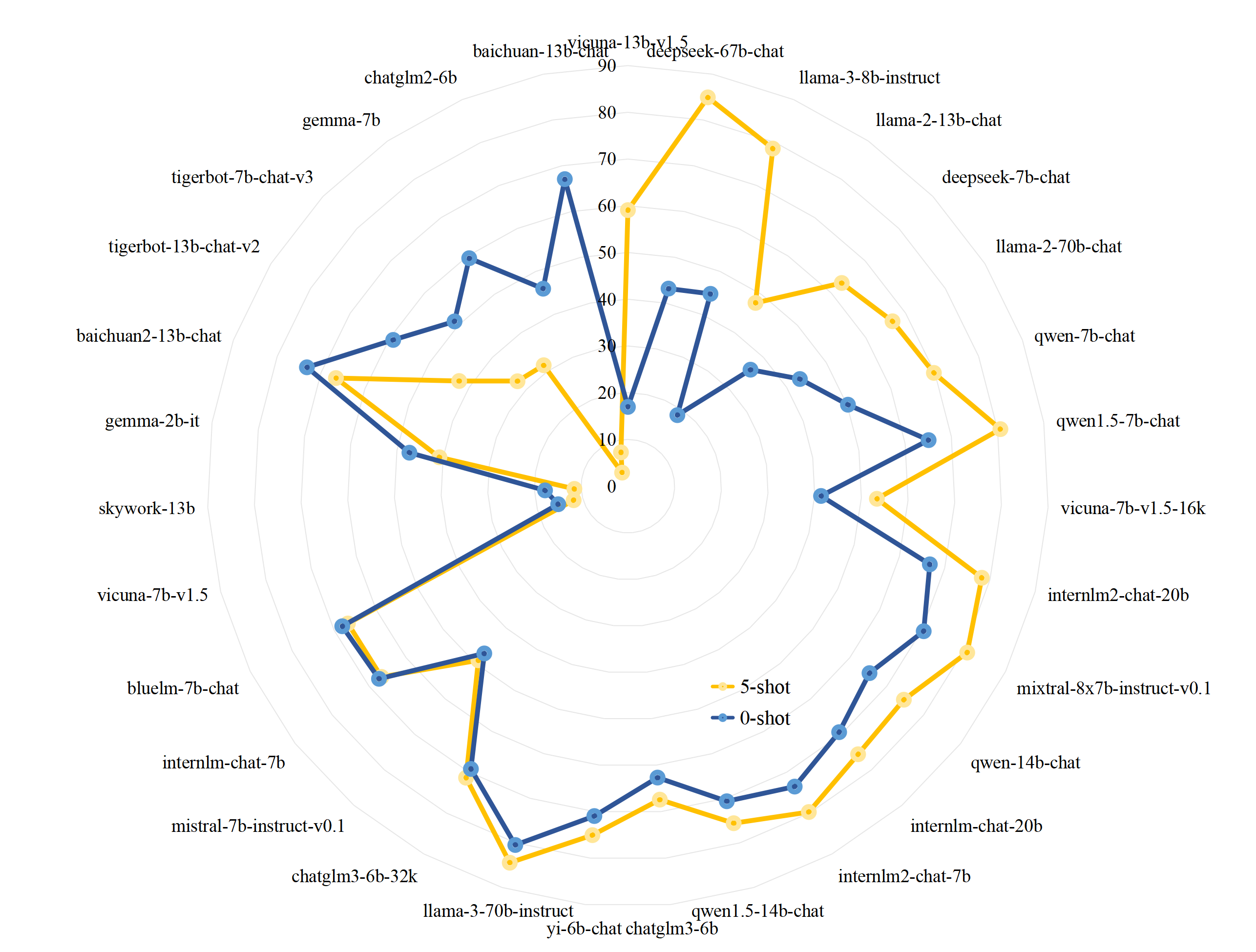}
    \caption{A radar chart of the composite index scores $I$ for 31 large language models evaluated in 5-shot and 0-shot settings on enviroexam.}
    \label{fig:radar-chart}
\end{figure}

\section{Discussion}
This study establishes a multi-source evaluation standard for large language models in the field of environmental science. By incorporating the coefficient of variation as an indicator, it evaluates the performance of mainstream open-source large language models in the domain of environmental science from multiple sources. This can provide an effective set of selection criteria for choosing large language models for fine-tuning in the field of environmental science.

Future research will follow the phi-3 approach, using specialized textbooks in environmental science to construct a more domain-specific dataset with a distinct environmental science focus.

\section*{Limitations}

Most of our data drafts come from GPT-4 and Claude. Given that existing large language models have been extensively trained on vast amounts of internet data, there is a risk of test data leakage. These models may have been trained on similar task formats or even the exact test data, resulting in exceptionally high scores~\cite{schaeffer2023pretraining}. In the future, we will seek more principled methods to prevent data contamination.

\section*{Acknowledgments}
We utilized 4 NVIDIA A100-80G GPUs for our tests. Given that the largest model is only 70B, most computations were completed within a few hours. This research was supported by the high-performance computing platform at Harbin Institute of Technology.

%
%
%
%
%
%
%
%
%
%
%
%
%
%
%
%
%
%

%

\clearpage
\appendix
\section{List of Environmental Science Core Classes}
\label{appendixA}

\begin{table}[h!]
\centering
\begin{tabular}{l l r}
\toprule
\textbf{Course Name} & \textbf{Education Plan} & \textbf{Numbers} \\ 
\midrule
Environmental Impact Assessment & Bachelor & 29 \\ 
Atmospheric Pollution Control Engineering & Bachelor & 26 \\ 
Environmental Monitoring & Bachelor & 27 \\ 
Pollution Control Microbiology & Bachelor & 23 \\ 
Environmental Nanotechnology & Bachelor & 19 \\ 
Principles And Applications Of Ecology & Bachelor & 19 \\ 
Environmental Physical Pollution And Control & Bachelor & 25 \\ 
Environmental Planning And Management & Bachelor & 25 \\ 
Environmental Analytical Chemistry & Bachelor & 11 \\ 
Environmental Science & Bachelor & 23 \\ 
Solid Waste Treatment And Resource Utilization & Bachelor & 23 \\ 
Environmental Soil Science & Bachelor & 14 \\ 
Specialty Wastewater Treatment & Bachelor & 14 \\ 
Environmental Decision Support System & Bachelor & 23 \\ 
Environmental Statistics & Bachelor & 24 \\ 
Environmental Economics & Bachelor & 24 \\ 
Indoor Environmental Pollution And Control & Bachelor & 28 \\ 
Industrial Ecology & Bachelor & 27 \\ 
Urban Water Supply And Drainage System Engineering & Bachelor & 14 \\ 
Environmental Fluid Mechanics & Bachelor & 11 \\ 
Water Pollution Control Engineering & Bachelor & 13 \\ 
Environmental Geographic Information System & Master & 29 \\ 
Modern Biology & Master & 23 \\ 
Atmospheric Particulate Matter Detection And Analysis Techniques & Master & 25 \\ 
Environmental Microwave Chemistry Technology & Master & 19 \\ 
Circulate Economy And Industrial Ecology Methodologies & Master & 27 \\ 
Clean Production And Energy Conservation And Emissions Reduction & Master & 22 \\ 
Reaction Kinetics And Reactor Design & Master & 15 \\ 
Engineering Ethics And Environmental Ethics & Master & 28 \\ 
Safe Disposal And Resource Recovery Of Hazardous Waste & Master & 23 \\ 
Sludge Treatment Disposal And Resource Utilization & Master & 19 \\ 
Theory And Technology Of Air Pollution Prevention And Control & Master & 22 \\ 
Theory And Technology Of Environmental Pollution Prevention And Control & Master & 21 \\ 
Environmental Electrochemical Theory And Technology & Master & 18 \\ 
Advanced Environmental Chemistry & Doctor & 25 \\ 
Modern Detection Techniques & Doctor & 34 \\ 
Persistent Organic Pollutants & Doctor & 26 \\ 
Carbon Neutrality Theory And Technology & Doctor & 27 \\ 
Environmental Pollution Investigation And Tracing & Doctor & 28 \\ 
Environmental Behavior Of Atmospheric Science And Climate Change & Doctor & 22 \\ 
Environmental Biotechnology & Doctor & 20 \\ 
Mass Transfer Processes & Doctor & 21 \\ 
\bottomrule
\end{tabular}
\caption{Environmental science education plan \& number of questions}
\end{table}

\clearpage
\section{Prompts used for each subject}
\label{appendixB}
\begin{figure}[h]
    \centering
    \includegraphics[width=1\linewidth]{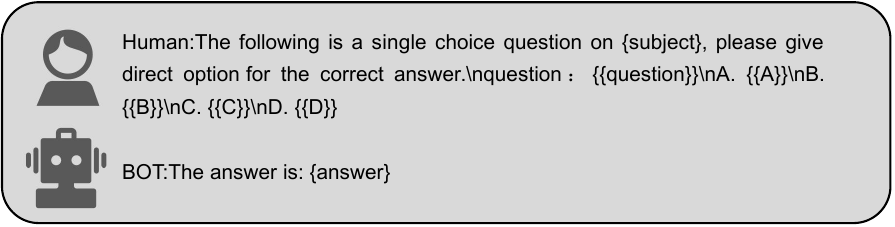}
    \caption{Prompts for each subject(Document translated, original question in Chinese)}
    \label{fig:enter-label}
\end{figure}

\section{Examples of questions}
\label{appendixC}
\begin{figure}[h]
    \centering
    \includegraphics[width=1\linewidth]{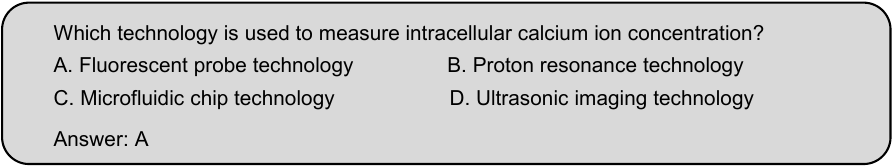}
    \caption{Examples from Environmental Electrochemistry Theory and Technology (Document translated, original question in Chinese)}
    \label{fig:enter-label}
\end{figure}

\section{Detailed Model Descriptions}
\label{appendixD}
\textbf{Baichuan2-13b-chat} and \textbf{Baichuan-13b-chat} are the new generation open-source large language models released by Baichuan Intelligent. The Baichuan2 series models are trained on high-quality corpora and have achieved the best results of their size on authoritative Chinese and English benchmarks. The models have been extensively tested on authoritative datasets in six domains: general, legal, medical, mathematics, code, and multilingual translation, all yielding favorable test results.

\textbf{ChatGLM2-6B} is the second-generation version of the open-source bilingual dialogue model ChatGLM-6B. Building on the development experience of the initial ChatGLM model, it has comprehensively upgraded the base model, achieving significant performance improvements across various datasets.

\textbf{ChatGLM3-6B} is the latest generation of the ChatGLM series open-source models. Its base model, ChatGLM3-6B-Base, utilizes more diverse training data and more rational training strategies. Evaluations on datasets from different angles such as semantics, mathematics, and knowledge indicate that ChatGLM3-6B-Base has the strongest performance among pre-trained models under 10B.

\textbf{ChatGLM3-6B-32K} further enhances the understanding of long texts based on ChatGLM3-6B. By updating position encoding and adopting more targeted long-text training methods, it achieves better handling of 32K-length contexts.

\textbf{DeepSeek-7B-Chat} and \textbf{DeepSeek-67B-Chat} are advanced language models from the DeepSeek LLM series. They have been trained from scratch on a massive dataset comprising 2 trillion English and Chinese tokens.

\textbf{Gemma-7B} and \textbf{Gemma-2B-it} are lightweight, state-of-the-art open models introduced by Google in the Gemma series. The Gemma models are well-suited for various text generation tasks, including Q\&A, summarization, and reasoning. They are built using the same research and technology used to create the Gemini models. Their relatively small size allows users to deploy these models in resource-limited environments.

\textbf{InternLM-Chat-20B} and \textbf{InternLM-Chat-7B} are large language models from the InternLM series. InternLM2-Chat-20B and InternLM2-Chat-7B are models from the InternLM2 series. These models are chat-oriented and tailored for practical scenarios. Compared to the InternLM series, the InternLM2 series models significantly outperform their predecessors in all aspects, particularly in reasoning, mathematics, coding, chat experience, instruction following, and creative writing, leading in performance among similarly sized open-source models, and supporting a wider range of intelligent agents and complex task multi-step tool calls.

\textbf{Mistral-7B-Instruct-v0.1} and \textbf{Mixtral-8x7B-Instruct-v0.1} are large language models from the Mistral series. They outperform the Llama 2 13B model in most benchmarks, approach CodeLlama 7B performance in coding tasks, while maintaining good performance in English tasks, and use Grouped Query Attention (GQA) to speed up inference.

\textbf{Qwen1.5-14B-Chat} and \textbf{Qwen1.5-7B-Chat} are test versions of large language models from the Qwen2 series. Qwen-14B-Chat and Qwen-7B-Chat are large language models from the Qwen series. These models are Transformer-based and pre-trained on large datasets. Compared to the Qwen series, the Qwen2 series models can stably support a 32K context length, support basic and chat models in multiple languages, and excel in language understanding, reasoning, and mathematics.

\textbf{Vicuna-13B-v1.5}, \textbf{Vicuna-7B-v1.5}, and \textbf{Vicuna-7B-v1.5-16K} are large language models from the Vicuna series, fine-tuned on user-shared conversations collected from ShareGPT. These models are primarily used for research on large language models and chatbots.

\textbf{BlueLM-7B-Chat} is a large-scale pre-trained language model independently developed by the Vivo AI Global Research Institute, open-sourcing a long-text base model and dialogue model supporting 32K. It was trained on a high-quality corpus with a scale of 26 trillion tokens, including Chinese, English, and a small amount of Japanese and Korean data, achieving leading results on C-Eval and CMMLU.

\textbf{TigerBot-13B-Chat-V2} and \textbf{TigerBot-7B-Chat-V3} are large language models from the TigerBot series. Developed on the basis of Llama-2 and BLOOM, the TigerBot series models further push the boundaries in data, training algorithms, infrastructure, and application tools. Compared to SOTA open-source models like Llama-2, our models achieve significant performance improvements, especially with a 6\% increase in English and a 20\% increase in Chinese. Additionally, the TigerBot model series has achieved leading performance on major academic and industrial benchmarks and leaderboards.

\textbf{Llama-2-70B-Chat} and \textbf{Llama-2-13B-Chat} are large language models from the Llama2 series developed by Meta. Llama-3-8B-Instruct and Llama-3-70B-Instruct are large language models from the Llama3 series developed by Meta. These models are optimized for dialogue use cases, outperforming many available open-source chat models on common industry benchmarks and optimized for usability and safety.

\textbf{Skywork-13B} is a large language model from the Skywork series. It is pre-trained on a high-quality, clean-filtered dataset of 3.2 trillion multilingual (mainly Chinese and English) and code data, with English accounting for 52.2\%, Chinese for 39.6\%, and code for 8\%. It demonstrates the best performance among models of the same scale in various evaluations and benchmarks while ensuring strong capabilities in both Chinese and English and code proficiency.

\textbf{Yi-6B-Chat} is a large language model from the Yi series. Targeting bilingual language models and a 3T multilingual corpus, the Yi series models have become one of the strongest LLM models globally, showing great prospects in language understanding, common sense reasoning, and reading comprehension.

\clearpage

\begin{landscape}
\section{Details of 5-shot and 0-shot evaluations for 31 LLMs}
\begin{figure}[h]
    \centering
    \includegraphics[width=1\linewidth]{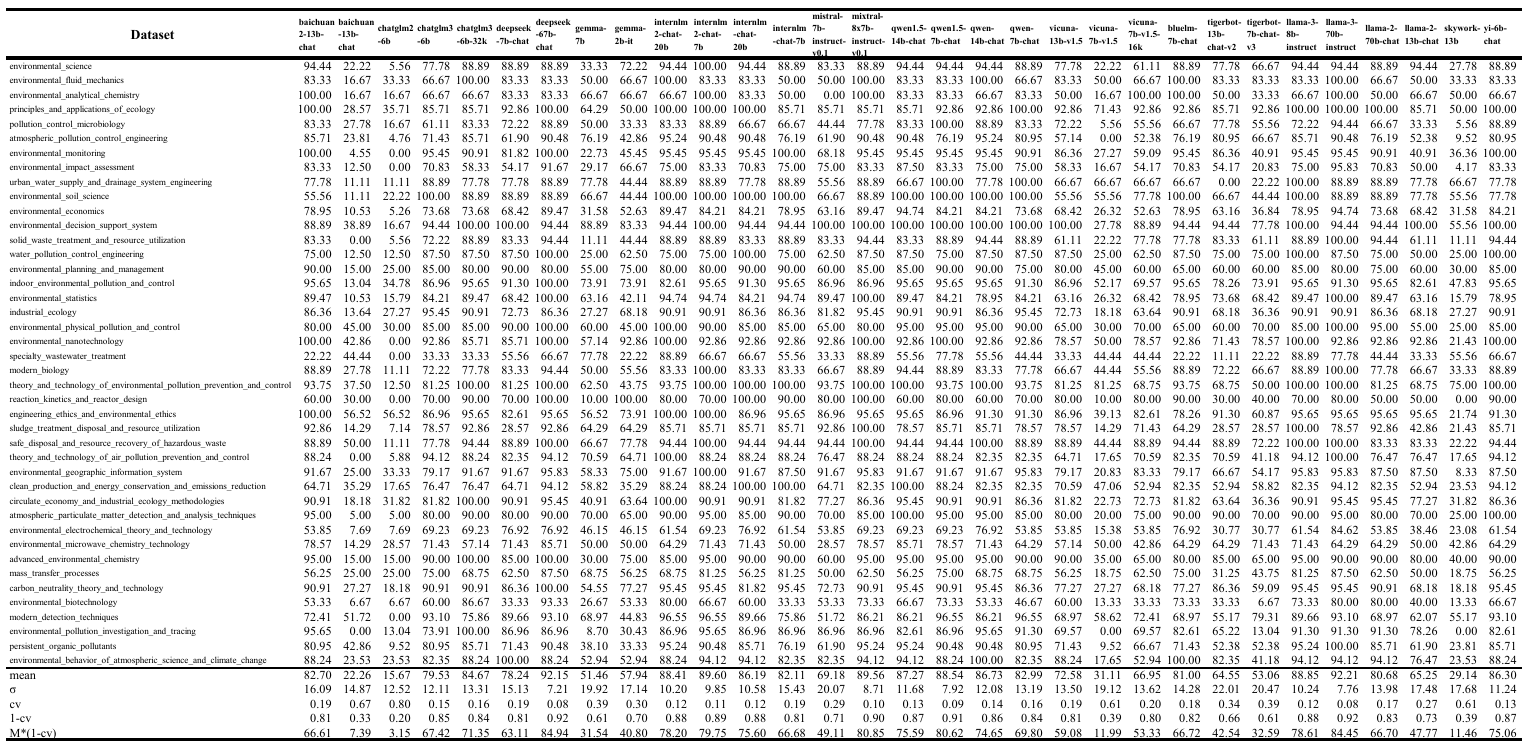}
    \caption{Detail scores for each subject in 5-shot}
    \label{fig:enter-label}
\end{figure}
\end{landscape}

\begin{landscape}
\begin{figure}[h]
    \centering
    \includegraphics[width=1\linewidth]{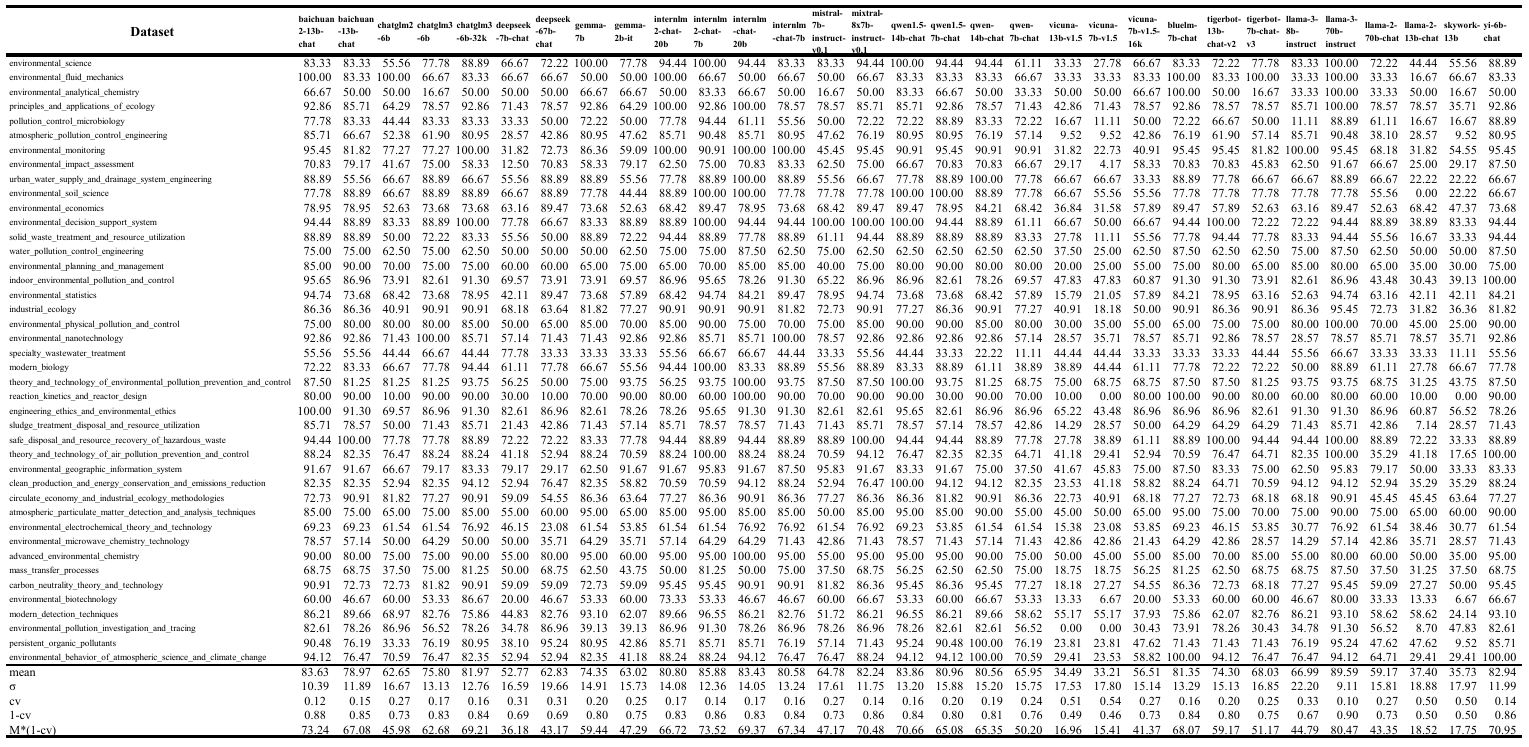}
    \caption{Detail scores for each subject in 0-shot}
    \label{fig:enter-label}
\end{figure}
\end{landscape}

\end{document}